\documentclass[10pt,onecolumn,a4paper]{article}

\usepackage{arxiv}

\usepackage[utf8]{inputenc} 
\usepackage[T1]{fontenc}    
\usepackage{hyperref}       
\usepackage{url}            
\usepackage{booktabs}       
\usepackage{amsfonts}       
\usepackage{nicefrac}       
\usepackage{microtype}      
\usepackage{lipsum}
\usepackage{float}
\usepackage{graphicx}
\usepackage{tikz}

\newcommand{\etal}{\textit{et al.}}

\title{Deep Classification Network for Monocular Depth Estimation}

\author{
  Azeez Oluwafemi\\
  Department of Electrical and Computer Engineering\\
  Carnegie Mellon University Africa, Kigali, Rwanda\\
  \texttt{oazeez@andrew.cmu.edu} \\
  \And 
  Yang Zou\\
  Department of Electrical and Computer Engineering\\
  Carnegie Mellon University, Pittsburgh, PA 15213\\
  \texttt{yzou2@andrew.cmu.edu} \\
  \And 
  B.V.K. Vijaya Kumar\\
  Department of Electrical and Computer Engineering\\
  Carnegie Mellon University Africa, Kigali, Rwanda\\
  \texttt{vk16@andrew.cmu.edu} \\
}

\begin{document}
\maketitle

\begin{abstract}
Monocular Depth Estimation is usually treated as a supervised and regression problem when it actually is very similar to semantic segmentation task since they both are fundamentally pixel-level classification tasks. We applied depth increments that increases with depth in discretizing depth values and then applied Deeplab v2 \cite{chen2018deeplab} and the result was higher accuracy. We were able to achieve a state-of-the-art result on the KITTI dataset\cite{geiger2013vision} and outperformed existing architecture by an 8\% margin.
\end{abstract}


\section{Introduction}

Generally trying to get 3D information from 2D images can be very challenging. This is because no exact solution exists, many 3D images could have produced the same 2D information. The task can feel like trying to create information from nothing. A task that is like that is Monocular depth estimation. It is a task that involves estimating depth from a single image rather than stereo pairs, which is what we're investigating in this paper. 

Depth estimation is an easier problem to track using stereo pairs \cite{scharstein2002taxonomy}. However, it is easy to now see it as a supervised learning challenge since ground truth depth maps could be obtained from stereo pairs and used to train models and predict depths \cite{saxena2006learning}. So deep convolutional networks can be used for building large networks that can predict depth maps \cite{eigen2015predicting}.

Our proposal is that depth estimation can be formulated as a pixel-level classification task similar to the semantic segmentation task. So a state-of-the-art architecture already performing well in semantic segmentation can be used for depth estimation. Depth values could be discretized so they have pixel-level classes just like in segmentation. We, therefore, applied spatially increasing discretization (SID) on the depth values. SID converts depth values to the logarithm scale, divides the range of values equally and then converts it back. 

The remaining sections in this paper are organized as follows. Relevant literature is first reviewed in Sec. \ref{Related Work}, then the proposed method is discussed in Sec. \ref{Methods}. In Sec. \ref{Experiments} We explore the performance of our proposed method through experiments and we finally conclude in Sec. \ref{Conclusion}

\section{Related Work} \label{Related Work}

\textbf{Depth Estimation} usually involves trying to estimate dense or sparse depth map given images. A common approach usually involves using stereo images. Points of correspondence are located in the stereo pairs and then depth is estimated by triangulation. \cite{scharstein2002taxonomy} did a taxonomy of stereo correspondence algorithms by comparing the performance of existing algorithms. It is also possible to learn depth estimation if it's posed as a supervised learning problem \cite{saxena2006learning}. The labels are usually monocular cues in 2D images \cite{hoiem2007recovering} \cite{ladicky2014pulling} \cite{li2014dept}. Markov Random Fields (MRFs) is usually used for getting global cues since some of the former features were merely local cues. MRF and supervised learning was used in Make3D \cite{saxena2009make3d}.
\cite{karsch2014depth} also gets global cues using Depth Transfer, which involves finding similar image in an already existing databases of images to the input image, then warping the candidate image and depth to align with the input image and finally using an optimization procedure to interpolate and smoothen the candidate depth values.

Using deep convolutional networks like VGG \cite{simonyan2014very} and ResNet \cite{he2016deep}, depth estimation performance has improved. \cite{eigen2015predicting} used a multi-scale network to predict depth, estimate surface normal and for semantic segmentation. 
\cite{fu2018deep} discretized depth with spatially increasing discretization (SID) and recast depth estimation as an ordinal regression problem.

\textbf{Semantic Segmentation} is a task where the aim is to predict the class each pixel of an image belongs to. A popular approach in this area is the use of a fully convolutional network (FCN) for prediction \cite{wu2015fully}. A much more advanced approach is Deeplab v2 \cite{chen2018deeplab}. Input images were passed through a layer of atrous (dilated) convolutional network which helps with adjusting the field of view and control the resolution of the feature map generated by a deep convolutional neural network. A coarse score map is then produced which undergoes bilinear interpolation for upsampling and finally fully connected conditional random field (CRF) is used as a post-processing process which helps incorporate low-level details in the segmentation result since skip connections are not used here. In an attempt to close domain gaps, the main idea has been to reduce the gap between source and target distribution by learning embeddings that are invariant to domains such as in Deep Adaptation Network (DAN) architecture which generalizes simple convolutional network tasks to fit for domain adaptation \cite{long2015learning}. The base convolutional neural network for Deeplab v2 is ResNet101 \cite{he2016deep} which has been pretrained on Imagenet dataset \cite{deng2009imagenet} through transfer learning technique \cite{raina2007self}.

\textbf{Domain Adaptation} involves trying to learn from a source distribution and predict on a target distribution. It is unsupervised when the source is labeled while the target is not. Deep Adaptation Network (DAN) tries to learn transferable features in task-specific layers which help generalize CNN for domain adaptation \cite{pmlr-v37-long15}. Deep CORAL can help unsupervised domain adaptation by aligning correlation of activation layers \cite{sun2016deep}. 

\textbf{Adaptation for Depth Estimation} involves reducing the domain gap between a source distribution (usually a synthetic dataset) and a target distribution (usually a real dataset) while trying to accomplish the task of unsupervised depth estimation in the unlabeled target domain. There are some important recent works in this aspect. AdaDepth (Adaptation for depth estimation) uses a residual encoder-decoder architecture with adversarial set up to accomplish a pixel-wise regression task of monocular depth estimation \cite{nath2018adadepth}. \cite{atapour2018real} accomplishes the same task by combining adversarial techniques and style transfer. GASDA (Geometry-Aware Symmetric Domain Adaptation Network) is an architecture that exploits epipolar geometry in the target domain and labels in the source domain\cite{zhao2019geometry}.

\section{Methods} \label{Methods}

This section first introduces the base architecture which was originally designed and is usually used for semantic segmentation and then how it was adapted for depth estimation by discretizing continuous depth ground truth. Each depth range is then treated as a class for semantic segmentation.

\subsection{Deeplab for Semantic Segmentation}

The base network used for this project is Deeplab v2. Prior to Deeplab, Fully connected convolutional networks were used for semantic segmentation \cite{long2015fully}. They had a prominent feature which was skip-connections, which involves feeding the output of one layer as an input to another layer skipping some layers in between. FCN-8 has 2 skip connections, FCN-16 has 1 and FCN-32 has none. The skip connections help with incorporating low-level details to the results of semantic segmentation. Deeplab v2 had some important features as shown in Figure \ref{fig:deeplab}. 

\begin{figure}[H]
\begin{center}
\includegraphics[width=1\linewidth]
                   {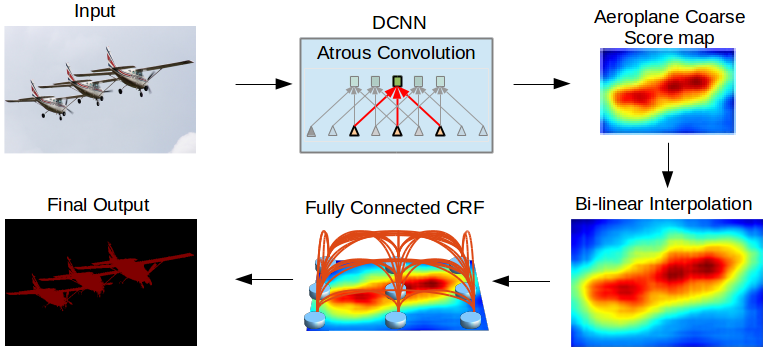}
\end{center}
   \caption{\textbf{Deeplab \cite{chen2018deeplab}}. An input image is passed through a Deep Convolutional Neural Network(DCNN) such as ResNet101, using atrous convolution to reduce downsampling. The score map output is then interpolated for upsampling to original image resolution. Low-level details are finally incorporated with the final result through a pre-processing step of fully connected conditional random field (CRF)}
\label{fig:Deeplab}
\end{figure}

Atrous convolution helps with adjusting the field of view and controlling the resolution of the feature map generated by a deep convolutional neural network.

Bilinear interpolation is used for upsampling and Fully connected CRF is done as a post-processing process which helps incorporated low-level details in the segmentation result since skip connections are not used here.

ASPP (Atrous spatial pyramid pooling) helps handle scale variability in semantic segmentation by adjusting the resolution of feature maps by applying various atrous rates and fusing the result.

\subsection{Spatially-Increasing Discretization}

\begin{figure}[H]
\centering
\begin{tikzpicture}
\begin{scope}
    \draw[|-|] (0,0) -- (10,0);
\node [below] at (0,0) {$\alpha$};
\node [below] at (10,0) {$\beta$};
\end{scope}

\begin{scope}[yshift=-1cm]
    \draw[|-|] (0,0) -- (10,0);
\node [below] at (0,0) {$d_{0}$};
\node [below] at (10,0) {$d_{K}$};
\foreach \x in {2,4,6,8}
    {\pgfmathtruncatemacro{\tmp}{0.5*\x}
    \draw ( \x,0.1) -- (\x,-0.1);
    \node [below] at (\x, -0.1 ) {$d_{\tmp}$};}
\end{scope}

\begin{scope}[yshift=-2cm]
    \draw[|-|] (0,0) -- (10,0);
\node [below] at (0,0) {$d_{0}$};
\node [below] at (10,0) {$d_{K}$};
\foreach \x [count = \xi] in {1,2,4,8}
    {\draw ( \x,0.1) -- (\x,-0.1);
    \node [below] at (\x, -0.1 ) {$d_{\xi}$};}
\end{scope}

\end{tikzpicture}
\caption{\textbf{Intervals.} UD (middle) and SID (bottom) to discretize depth interval [$\alpha, \beta$]} into sub intervals $d_i$ where $i = 0, 1,2,...K$ and K = number of class. UD is Uniform Discretization, SID is spatially increasing interval discretization \label{fig:SID}
\end{figure}
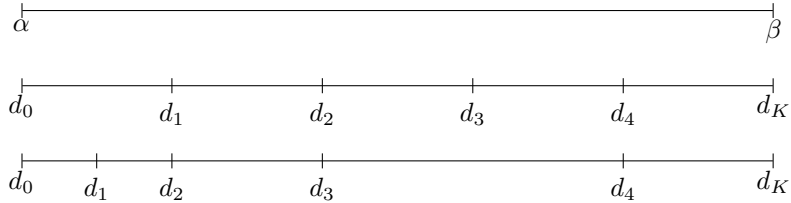

In order to properly frame depth estimation as a pixel-level classification problem which is exactly how semantic segmentation is done, then there's a need to discretize continuous depth values. This is done such that each bin of depth values can be treated as a class of its own. Two common methods used to discretize continuous depth values are uniform discretization (UD) and spatially increasing interval discretization (SID). Uniform discretization just divides the space between the interval [$\alpha, \beta$] equally as shown in Fig \ref{fig:SID}. To divide the interval [$\alpha, \beta$] into K classes, UD can be formulated as:
\begin{equation}
    d_i = \alpha + (\beta - \alpha) * \frac{i}{K}, i = 0, 1, ..., K
\end{equation}
where $ d_i \in {d_0, d_1,..., d_K}$ are depth interval boundaries. One challenge of using UD for discretization is that large depth values like the pixels belonging to the class "sky" will correspond to too many intervals or equivalently too many values. They could become really noisy and influence the loss. One of the common ways of handling outliers is to apply a log function to them. This means log function is applied to the continuous depth values and then divided equally in the interval [$\alpha, \beta$]. This is exactly what SID addresses. We still however treated depth values larger than 80m. Using SID to divide the interval [$\alpha, \beta$] into K classes, we can formulate it as:
\begin{equation}
    d_i = \exp(\log{\alpha} + \frac{\log{\beta/\alpha} * i}{K}), i=0,...,K
\end{equation}
where $ d_i \in {d_0, d_1,..., d_K}$ are depth interval boundaries.
\section{Experiments} \label{Experiments}
\vspace{-5mm}

We now discuss the implementation details of our experiment, and our evaluation on two datasets KITTI \cite{geiger2013vision} and SYNTHIA \cite{ros2016synthia}. 

\textbf{Implementation details} We used the publicly available deep learning framework pytorch \cite{paszke2017pytorch}. Using ResNet-101 \cite{he2016deep} as our feature extractor with Imagenet-pretrained weights, our model is trained for 100K iterations for KITTI and SYNTHIA with a batch size of 3 and initial learning rate of 0.00025 which applies a polynomial decay with the power of 0.9. We used weight decay of 0.0005 and momentum of 0.9. Eigen test split, the data augmentation techniques from \cite{eigen2014depth} was used. Our model was trained on a 12gb RAM NVIDIA TITAN Xp GPU.

\begin{table}[H]
  \begin{center}
    \begin{tabular}{@{}rrrrcrrrcrrr@{}}\toprule
    & \multicolumn{3}{c}{\textbf{Higher is better}} & \multicolumn{3}{c}{\textbf{Lower is better}} \\
    \cmidrule(r{1.0ex}){2-4}\cmidrule(l{1.0ex}){5-7}
      \textbf{Method} & $\delta < 1.25$ & $\delta < 1.25^2$ & $\delta < 1.25^3$ & Abs Rel & RMSE & $RMSE_{log}$\\
      \midrule
      Make3D\cite{saxena2007learning} & 0.601 & 0.820 & 0.926 & 0.280 & 8.734 & 0.361\\
      Eigen \etal\cite{eigen2014depth} & 0.692 & 0.899 & 0.967 & 0.190 & 7.156 & 0.270 \\
      Liu \etal\cite{liu2016learning} & 0.647 & 0.882 & 0.961 & 0.217  & 6.986 & 0.289\\
      LRC (CS + K)\cite{godard2017unsupervised} & 0.861 & .949 & 0.976 & 0.114 & 4.935 & 0.206\\
      Kuznietsov \etal\cite{kuznietsov2017semi} & 0.862 & 0.960 & 0.986 & .113 & 4.621 & 0.189\\
      DORN (VGG)\cite{fu2018deep} & 0.915 & 0.980 & 0.993 & 0.081 & 3.056 & 0.132\\
      DORN (ResNet)\cite{fu2018deep} & \textbf{0.932} & 0.984 & 0.994 & \textbf{0.072} & 2.727 & \textbf{0.120}\\
      \textbf{Ours(ResNet)} & 0.796 & \textbf{0.985} & \textbf{1.000} & 0.075 & \textbf{2.499} & 0.156\\
      \bottomrule
    \end{tabular}
     \caption{Performance on KITTI. K is KITTI, CS is Cityscapes. $1.25$, $1.25^2$ and $1.25^3$ are pre-defined thresholds for Accuracy under threshold metric ($\delta$)}
    \label{tab:table1}
  \end{center}
\end{table}

\begin{table}[H]
  \begin{center}
    \begin{tabular}{@{}rrrrcrrrcrrr@{}}\toprule
    & \multicolumn{3}{c}{\textbf{Higher is better}} & \multicolumn{3}{c}{\textbf{Lower is better}} \\
    \cmidrule(r{1.0ex}){2-4}\cmidrule(l{1.0ex}){5-7}
      \textbf{Method} & $\delta < 1.25$ & $\delta < 1.25^2$ & $\delta < 1.25^3$ & Abs Rel & RMSE & $RMSE_{log}$\\
      \midrule
      \textbf{Ours(ResNet)} & 0.785 & 0.937 & 0.945 & 0.108 & 9.270 & 0.206\\
      \bottomrule
    \end{tabular}
     \caption{Performance on SYNTHIA.$1.25$, $1.25^2$ and $1.25^3$ are pre-defined thresholds for Accuracy under threshold metric ($\delta$)}
    \label{tab:table2}
  \end{center}
\end{table}

\begin{figure}[H]
\begin{center}
\includegraphics[width=0.8\linewidth]
                   {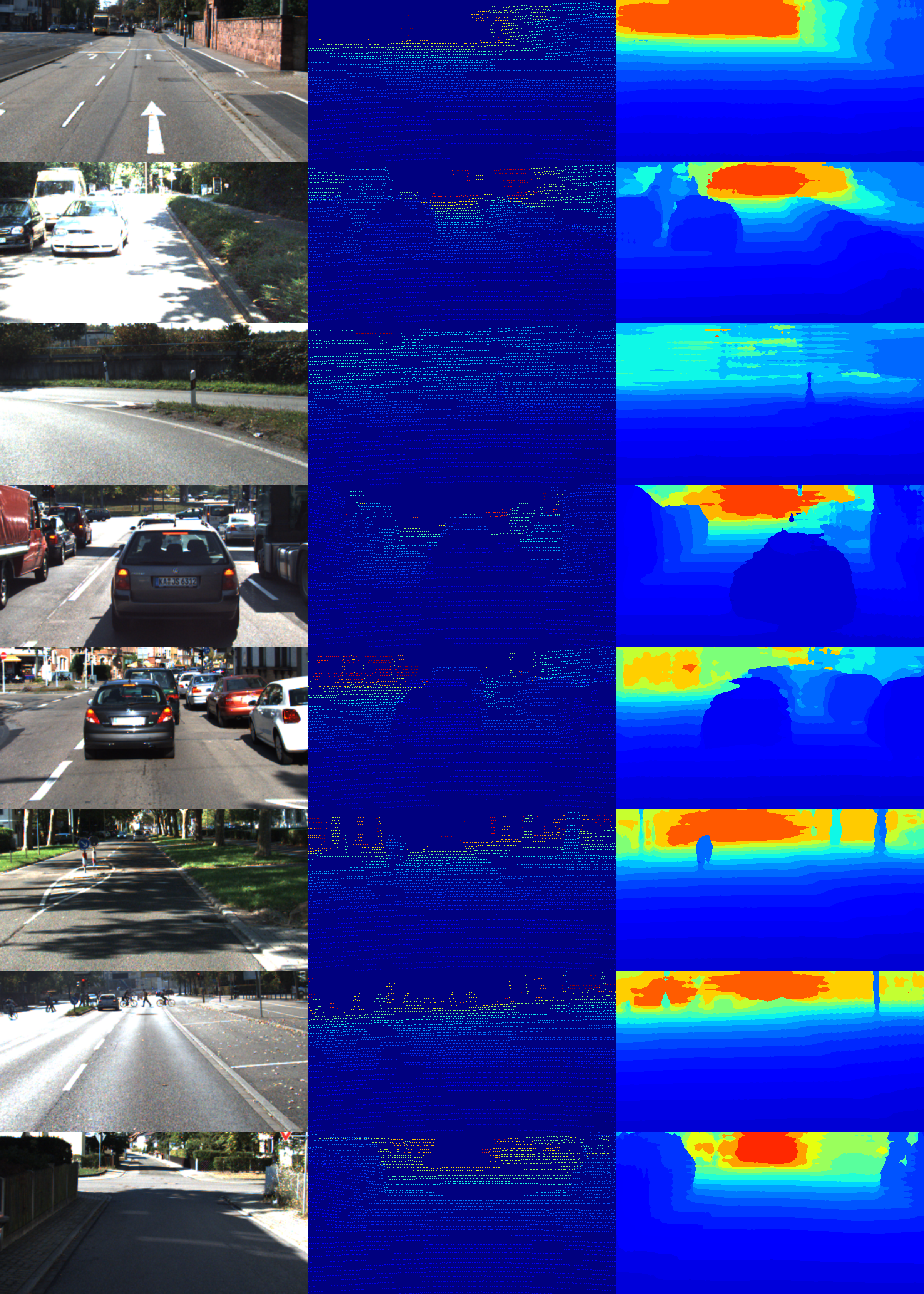}
\includegraphics[width=0.8\linewidth]
                   {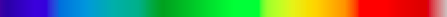}                   
\end{center}
   \caption{\textbf{Depth Predictions on KITTI}. The Image on the left column, ground truth in the middle and our model prediction on the right. The color color bar below shows the range of depths. Blue is nearer.}
\label{fig:kitti}
\end{figure}

\begin{figure}[H]
\begin{center}
\includegraphics[width=0.8\linewidth]
                   {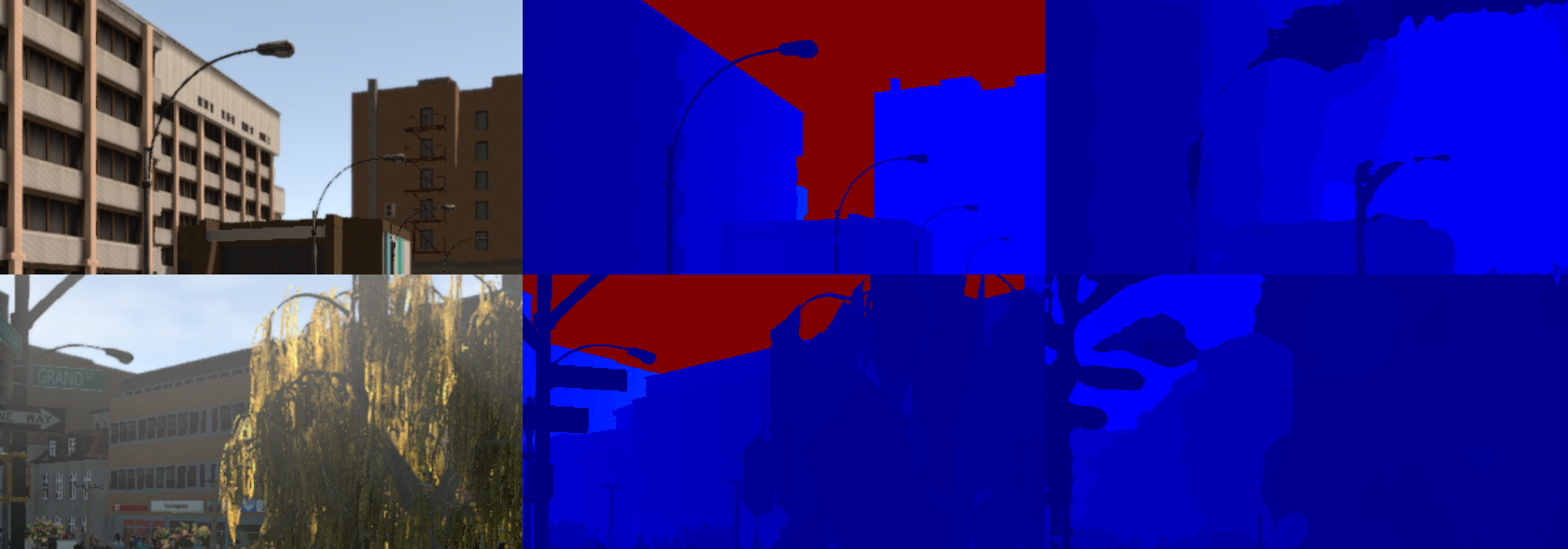}
\includegraphics[width=0.8\linewidth]
                   {color.png}
\end{center}
   \caption{\textbf{Depth Predictions on SYNTHIA}. The Image on the left column, ground truth in the middle and our model prediction on the right. The color color bar below shows the range of depths. Blue is nearer.}
\label{fig:synthia}
\end{figure}

\subsection{Benchmark Performance}

\textbf{KITTI} The KITTI dataset \cite{geiger2013vision} contains images from cameras and depth sensors collected from a car. There are 61 total scenes. We used 23488 images from 32 scenes for training and 697 images from 29 scenes for testing \cite{eigen2014depth}. The images were cropped to 420 x 800. We used SID to discretize the depth maps into 71 classes and ignored depth values beyond 80m during network loss calculations. 

\textbf{SYNTHIA} We used the SYNTHIA dataset \cite{ros2016synthia} because it's synthetic. We used the SYNTHIA-RAND-CITYSCAPES subset which has 9,400 labeled images. We also cropped to 420 x 800 for training and testing. Depth map was also discretized to 71 classes and depth values beyond 80m were also ignored.

\textbf{Performance} Table \ref{tab:table1} shows the result of our performance compared to other existing models on KITTI. The metrics used for comparison are standards compared in \cite{cadena2016measuring}. Letting $\hat{d}_p$ and $d_p$ denote estimated and ground truth depths respectively at pixel $p$, T denote total number of pixels, the metrics used were calculated thus:

Absolute Relative Error \cite{saxena2009make3d},
\begin{equation}
    absRel = \frac{1}{T}\sum_p \frac{|d_p - \hat{d}_p|}{d_p} 
\end{equation}
Root Mean Square Error \cite{li2010towards},
\begin{equation}
    RMSE = \sqrt{\frac{1}{T} \sum_p (d_p - \hat{d}_p)^2}
\end{equation}
log scale invariant Root Mean Square Error \cite{eigen2014depth},
\begin{equation}
    RMSE_{log} = \frac{1}{T}\sum_p (\log{\hat{d}_p} - \log{d_p} + \alpha(\hat{d}_p, d_p) )^2
\end{equation}
where $\alpha(\hat{d}_p, d_p)$ addresses scale alignment.

Accuracy under a threshold \cite{ladicky2014pulling}
\begin{equation}
    \delta < th = max(\frac{\hat{d}_p}{d_p}, \frac{d_p}{\hat{d}_p})
\end{equation}
where th is a predefined threshold, we used $1.25$, $1.25^2$ and $1.25^3$

We were able to get $\sim 8\% $ improve in accuracy in terms of root mean squared error metrics. Our model also performed better in other metrics. Table \ref{tab:table2} also shows the result of our model's performance on SYNTHIA. Qualitative results are also shown in Figures \ref{fig:kitti} and \ref{fig:synthia}. The results demonstrate that our model is applicable to outdoor data and synthetic dataset.

\section{Conclusion} \label{Conclusion}

We've been able to demonstrate how a depth estimation task can be formulated as a semantic segmentation problem since the two tasks are fundamentally just pixel-level classification tasks at the core. Simply discretizing the depth map and treating the binned depths as classes and applying a state-of-the-art semantic segmentation network can produce a result that outperforms existing results. In the future, we would investigate domain gaps in monocular depth maps estimation and apply class balanced self training\cite{zou2018unsupervised} to attempt to reduce the gap.

\bibliographystyle{unsrt}  
\bibliography{main}  


\end{document}